\pdfoutput=1
\documentclass[conference]{IEEEtran}
\IEEEoverridecommandlockouts
\usepackage{amsmath,graphicx}

\usepackage[latin1]{inputenc}
\usepackage{amsthm, amsfonts, amssymb, mathrsfs, amsmath}
\usepackage{amsmath}
{
	\theoremstyle{plain}
	
}
\usepackage{nicefrac}
\usepackage{mathtools}
\usepackage{multirow}
\usepackage[ruled,linesnumbered]{algorithm2e}
\let\oldnl\nl
\newcommand{\nonl}{\renewcommand{\nl}{\let\nl\oldnl}}
\usepackage[mathcal]{euscript}
\usepackage[draft]{fixme}
\usepackage{enumerate}
\usepackage{dsfont}
\usepackage{tikz}
\usepackage{subfigure}
\allowdisplaybreaks[1]
\usepackage{scalerel}
\usepackage{mathrsfs}

\newcommand{\bM}{{\boldsymbol{\mathrm{M}}}}
\newcommand{\bP}{{\boldsymbol{\mathrm{P}}}}
\newcommand{\bQ}{{\boldsymbol{\mathrm{Q}}}}
\newcommand{\bH}{{\boldsymbol{\mathrm{H}}}}
\newcommand{\bI}{{\boldsymbol{\mathrm{I}}}}

\newcommand{\bL}{{\boldsymbol{\mathrm{L}}}}
\newcommand{\bS}{{\boldsymbol{\mathrm{S}}}}

\newcommand{\bW}{{\boldsymbol{\mathrm{W}}}}
\newcommand{\bU}{{\boldsymbol{\mathrm{U}}}}

\newcommand{\bX}{{\boldsymbol{\mathrm{X}}}}

\newcommand{\bs}{{\boldsymbol{\mathrm{s}}}}

\newcommand{\bq}{\boldsymbol{\mathrm{q}}}

\newcommand{\q}{\mathrm{q}}
\newcommand{\s}{\mathrm{s}}

\newcommand{\uu}{\mathrm{u}}

\def\BibTeX{{\rm B\kern-.05em{\sc i\kern-.025em b}\kern-.08em
  T\kern-.1667em\lower.7ex\hbox{E}\kern-.125emX}}
\begin{document}

\title{A Deep-Unfolded Reference-Based RPCA Network For Video Foreground-Background Separation\\
}



\author{\IEEEauthorblockN{Huynh Van Luong$^{\ast\dagger}$, Boris Joukovsky$^{\ast\dagger}$, Yonina C. Eldar$^{\ddagger}$, Nikos Deligiannis$^{\ast\dagger}$}
\vspace{1ex}
\IEEEauthorblockA{$^{\ast}$Department of Electronics and Informatics, Vrije Universiteit Brussel, Pleinlaan 2, B-1050 Brussels, Belgium \\
$^{\dagger}$imec, Kapeldreef 75, B-3001 Leuven, Belgium\\
$^{\ddagger}$Department of Mathematics and Computer Science, Weizmann Institute of Science, Rehovot 7610001, Israel\\
}
}




\maketitle

\begin{abstract}
Deep unfolded neural networks are designed by unrolling the iterations of optimization algorithms. They can be shown to achieve faster convergence and higher accuracy than their optimization counterparts. This paper proposes a new deep-unfolding-based network design for the problem of Robust Principal Component Analysis (RPCA) with application to video foreground-background separation. Unlike existing designs, our approach focuses on modeling the temporal correlation between the sparse representations of consecutive video frames. To this end, we perform the unfolding of an iterative algorithm for solving reweighted $\ell_1$-$\ell_1$ minimization; this unfolding leads to a different proximal operator (a.k.a. different activation function) adaptively learned per neuron. Experimentation using the moving MNIST dataset shows that the proposed network outperforms a recently proposed state-of-the-art RPCA network in the task of video foreground-background separation.

\end{abstract}


\section{Introduction}
\label{sec:intro}
Principal component analysis (PCA)~\cite{wold1987principal} has been a key method for data analysis with a plethora of applications in anomaly detection, dimensionality reduction, and signal compression among many. The solution of PCA---namely, the set of orthogonal basis vectors (the principal components) that define a subspace where the data lives---can be easily obtained by applying the singular vector decomposition (SVD) on a matrix $\bM$ formed by the data vectors. Robust PCA (RPCA)~\cite{candes2011robust} is a variant of PCA that addresses the sensitivity of SVD to outliers. RPCA decomposes the data matrix~$\bM$ into the sum of a low-rank component~$\bL$, whose subspace defines the principal components, and a sparse component $\bS$, which captures the outliers. 

RPCA has found various applications, including anomaly detection for networks~\cite{mardani2012dynamic} (e.g., computer networks and social media networks), reconstruction of dynamic magnetic resonance imaging (MRI)~\cite{otazo2015low}, and data visualization~\cite{slavakis2014modeling}. In video analysis, which is the domain this paper focuses on, RPCA has been used to decompose a sequence of vectorized frames, comprising the columns of $\bM$,
into the background modeled by the low-rank component $\bL$, and the foreground modeled by the sparse innovation component $\bS$. A similar decomposition was used in ultrasound to separate tissues from blood flow \cite{CohenICASSP19,solomon2019deep,SlounIEEE20}.

Several optimization-based methods have been proposed to solve the low-rank plus sparse matrix decomposition problem. Cand\'es \textit{et al.}~\cite{candes2011robust} suggested a convex formulation of the problem---referred to as principal component pursuit (PCP)---by using the $\ell_1$-norm and the nuclear-norm to encode the structure of the sparse and low-rank component, respectively. Furthermore, with the goal of achieving faster convergence, non-convex methods have been proposed based on alternating minimization~\cite{netrapalli2014non} and projected gradient descent~\cite{yi2016fast}. The study in~\cite{NarayanamurthyICASSP18} introduced a memory-efficient Robust PCA and its online version achieves nearly-optimal memory complexity. We refer to~\cite{vaswani2018robust} for a comprehensive overview.


Deep unfolding methods have been achieving state-of-the-art performance in solving decomposition problems in terms of both accuracy and computational complexity~\cite{GregorICML10,SreterICASSP18,LeICIP19,SprechmannPAMI15,CohenICASSP19,solomon2019deep,SlounIEEE20}. The study in~\cite{GregorICML10} proposed to unroll the iterations of the Iterative Shrinkage-Thresholding Algorithm (ISTA) to a feed-forward neural network---coined Learned ISTA (LISTA)---which is trained on data. The learned convolutional sparse coding network in~\cite{SreterICASSP18} and the deep-unfolded recurrent neural network in~\cite{LeICIP19} are convolutional and recurrent extensions of LISTA that solve the convolutional and the dynamic sparse representation problem, respectively. Regarding the low-rank-plus-sparse decomposition problem, the authors of \cite{SprechmannPAMI15} unrolled the iterations of proximal gradient methods to form a deep feed-forward neural network. Furthermore, the works~\cite{CohenICASSP19,solomon2019deep,SlounIEEE20} proposed the Deep Convolutional Robust PCA network (CORONA), which uses convolutional layers and an SVD for the low-rank approximation. 


In this paper, we present a deep unfolded RPCA network for the problem of video foreground-background separation. Our design aims to capture the inherent temporal correlation among the sparse representations of consecutive video frames. To this end, we propose a RPCA network that unfolds an iterative algorithm for solving reweighted $\ell_1$-$\ell_1$ minimization~\cite{Luong2018Compressive}, an extension of reweighted $\ell_1$ minimization~\cite{Candes08}. Our unfolded design leads to a new proximal operator with multiple thresholds (a.k.a. activation functions), which are adaptively learnt per neuron, thereby increasing network adaptivity and expressivity. Experimentation on the moving MNIST dataset~\cite{SrivastavaICML15} shows that the proposed network outperforms the CORONA~\cite{CohenICASSP19} network in terms of accuracy and convergence speed. 

The rest of the paper is as follows: Section~\ref{sec:background} presents the background and Section~\ref{sec:proposedNetwork} describes the proposed refRPCA network. The experimental evaluation of our model is given in Section~\ref{sec:experiment}, whereas Section~\ref{sec:conclusion} draws the conclusion. 

\section{Background on RPCA}
\label{sec:background}

In this section, we review existing optimization-based methods~\cite{JWright09, CandesRPCA,Venkat11} and deep-learning-based models~\cite{CohenICASSP19} addressing the RPCA problem.

\subsection{Optimization-Based Methods for RPCA} Traditional methods for RPCA~\cite{JWright09, CandesRPCA,Venkat11} decompose the data matrix~$\bM$ into~$\bS$ and $\bL$ by solving the principal component pursuit~(PCP) \cite{CandesRPCA}~problem:
\vspace{-2pt}
\begin{equation}\label{PCP}
\min_{\bL,\bS} \|\bL\|_{*}+\lambda\|\bS\|_{1} \text{~s.t.~}\bM=\bL+\bS,
\end{equation}
where $\|\bL\|_{*}=\sum_i\sigma_i(\bL)$ is the nuclear norm---sum of singular values~$\sigma_i(\bL)$---of the matrix~$\bL$, $\|\bS\|_1=\sum_{i,j}|\mathrm{s}_{i,j}|$ is the $\ell_1$-norm of~$\bS$ organized in a vector, and~$\lambda$ is a regularization parameter. The aforementioned RPCA methods~\cite{JWright09,CandesRPCA,Venkat11} typically assume that the~$\bL$ component lies in a low-dimensional subspace, i.e., the background frames are static or slowly-changing. In video foreground-background separation, a sequence of $m$ vectorized frames (modeled by~$\bM \in \mathbb{R}^{n\times m}$) is separated into the slowly-changing background~$\bL$ and the sparse foreground $\bS$. 

Problem~\eqref{PCP} can be formulated in a Lagrangian form
\begin{equation}\label{generalPCP}
\min_{\bL,\textcolor{black}{\bS}}\frac{1}{2}\|\bM-\bL-\bS\|^{2}_{F} + \lambda_1\|\bL\|_{*}+\lambda_2\|\bS\|_{1},
\end{equation}
where $\|\cdot\|_{F}$ denotes the Frobenious norm and $\lambda_1$ and $\lambda_2$ are tuning parameters. By using proximal gradient methods \cite{Beck09} to solve \eqref{generalPCP}, $\bL^{(k+1)}$ and $\bS^{(k+1)}$ at iteration $k\hspace{-1pt}+\hspace{-1pt}1$ can be iteratively computed via the singular value thresholding operator \cite{Cai10} for $\bL$ and the soft thresholding operator \cite{Beck09} for $\bS$.

\subsection{Deep Unfolding for RPCA}
The deep convolutional RPCA (CORONA) network in~\cite{solomon2019deep} considered measurement matrices $\bH_1$ and $\bH_2$ for the $\bL$ and $\bS$ components, respectively. The decomposition problem was then formulated as
\begin{equation}\label{corona}
\min_{\bL,\textcolor{black}{\bS}}\frac{1}{2}\|\bM-\bH_1\bL-\bH_2\bS\|^{2}_{F} + \lambda_1\|\bL\|_{*}+\lambda_2\|\bS\|_{1,2},
\end{equation}
where $\|\cdot\|_{1,2}$ is the mixed $\ell_{1,2}$ norm. Problem \eqref{corona} was solved via iteratively updating $\bL^{(k+1)}$ and $\bS^{(k+1)}$ at iteration $k+1$ with
\begin{subequations}
	\label{coronaProximal}
	\begin{align}
	\hspace{-3pt} \bL^{(k+1)}=&\varGamma_{\frac{\lambda_1}{c}}\Bigg(\hspace{-3pt}\Big(\bI-\frac{1}{c}\bH_1^T\bH_1\Big)\bL^{(k)}\hspace{-3pt}-\bH_1^T\bH_2\bS^{(k)}+\bH_1^T\bM\Bigg)\label{LProximal}\\
	\hspace{-3pt}\bS^{(k+1)}=& \varPhi_{\frac{\lambda_2}{c}}\Bigg(\hspace{-3pt}\Big(\bI-\frac{1}{c}\bH_2^T\bH_2\Big)\bS^{(k)} \hspace{-3pt}-\bH_2^T\bH_1\bL^{(k)}+\bH_2^T\bM\Bigg),\label{SProximal}
	\end{align}
\end{subequations}
where $\varGamma_{\frac{\lambda_1}{c}}(.)$ and $\varPhi_{\frac{\lambda_2}{c}}(.)$ are the singular value thresholding~\cite{Cai10} and mixed $\ell_{1,2}$ soft thresholding \cite{Beck09} operators, respectively, and $c$ is a Lipschitz constant.

Following the principles of deep unfolding, the authors of~\cite{CohenICASSP19} proposed to unroll the iterations of the algorithm solving Problem \eqref{corona} into a multiple-layer neural network, the $k^{th}$ layer of which computes:
\begin{subequations}
	\label{coronaLayer}
	\begin{align}
	\hspace{-3pt}\bL^{(k+1)}&\hspace{-3pt}=\varGamma_{\lambda_1^{(k)}}\Big\{\bW_1^{(k)}*\bM+\bW_3^{(k)}*\bS^{(k)}+\bW_5^{(k)}*\bL^{(k)}\Big\}\label{LLayer}\\
	\hspace{-3pt}\bS^{(k+1)}&\hspace{-3pt}= \varPhi_{\lambda_2^{(k)}}\Big\{\bW_2^{(k)}*\bM+\bW_4^{(k)}*\bS^{(k)} +\bW_6^{(k)}*\bL^{(k)}\Big\},\label{SLayer}
	\end{align}
\end{subequations}
where $*$ denotes the convolution operator. In CORONA, the weights of the convolutional layers $\bW_1^{(k)},\cdots,\bW_6^{(k)}$ and the regularization parameters $\lambda_1^{(k)}$, $\lambda_2^{(k)}$ are learned from training data using back-propagation.

\section{The Proposed refRPCA-Net Network}
\label{sec:proposedNetwork}

\subsection{The Proposed Unfolded Method}

We consider the problem of video foreground-background separation and attempt to capture the inherent temporal correlation in video. We assume that the foregrounds of two consecutive frames $\bs_{t-1},\bs_t$ in $\bS=[\bs_1,\cdots,\bs_m]$ are correlated via a projection matrix $\bP$, under the temporal correlation assumption, i.e., $\bs_t \approx \bP\bs_{t-1}$. For simplicity, $\bP$ is not varying across time. Then, we construct a sequence of reference frames $\bS_{\bP}$ by $\bS_{\bP} = [\bs_1,\bP\bs_1,\cdots,\bP\bs_{m-1}]$. We propose a reference-based RPCA (refRPCA) problem that leverages the correlated reference $\bS_{\bP}$ of the sparse component $\bS$ via $\ell_1$-$\ell_1$ minimization \cite{MotaTIT17,Luong2018Compressive} to improve the separation problem in \eqref{corona}. Furthermore, the proposed refRPCA uses a reweighting scheme \cite{Candes08, Luong2018Compressive} via reweighting the elements of $\bS$ with a matrix $\bQ$. The latter choice results from the observation that reweighted $\ell_1$-$\ell_1$-minimization \cite{Luong2018Compressive} outperforms its non-reweighted counterpart, leading to more accurate sparse representations. Thus, the refRPCA problem is formulated as 
\begin{align}\label{refRPCA}
\min_{\bL,\textcolor{black}{\bS}}&\frac{1}{2}\|\bM-\bH_1\bL-\bH_2\bS\|^{2}_{F} + \lambda_1\|\bL\|_{*}\nonumber\\
&+\lambda_2\|\bQ\circ\bS\|_{1}+\lambda_3\|\bQ\circ(\bS-\bS_{\bP})\|_{1},
\end{align}
where~\textquotedblleft$\circ$\textquotedblright~denotes element-wise multiplication and a weighting matrix $\bQ\in\mathbb{R}^{n\times m}$ is defined as $\bQ=[\bq,\cdots,\bq]$ with $\bq \in \mathbb{R}^{n}$, which consists of $m$ weighting vectors $\bq$.

We solve \eqref{refRPCA} using a proximal gradient method~\cite{Beck09}, where the low-rank component $\bL^{(k+1)}$ is iteratively computed via the singular value thresholding operator \cite{Cai10} as in \eqref{LProximal}. The sparse component $\bS^{(k+1)}$ is updated using a new proximal operator $\varPhi_{\frac{\lambda_2}{c},\frac{\lambda_3}{c},\bq,\bS_{\bP}}(.)$, which is formulated for the reweighted-$\ell_1$-$\ell_1$ minimization \cite{Candes08, Luong2018Compressive}, that is,
\begin{align}\label{proximalOperatorS}
\varPhi_{\frac{\lambda_2}{c},\frac{\lambda_3}{c},\bq,\bS_{\bP}}&(\bX)=\arg\min\limits_{\bU}\Big\{ \frac{\lambda_2}{c}\|\bQ\circ\bU\|_{1}\nonumber\\
&+\frac{\lambda_3}{c}\|\bQ\circ(\bU-\bS_{\bP})\|_{1} + \frac{1}{2}\|\bU-\bX\|^{2}_{2}\Big\}.
\end{align}


\begin{figure}[t]
	\centering 
	\subfigure[CORONA: $\varPhi_{\frac{\lambda_2}{c}}(\mathrm{x})$]{\label{proximalCORONA}\includegraphics[width=0.25\textwidth]{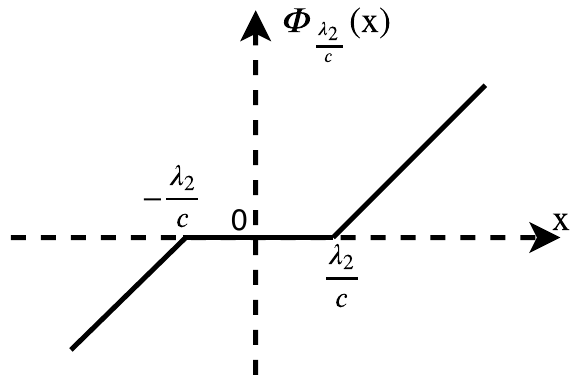}	}	
	\subfigure[refRPCA: $\varPhi_{\frac{\lambda_2}{c},\frac{\lambda_3}{c}\q,\s_{\bP}}(\mathrm{x})$ for $\s_{\bP}\geq 0$.]{\label{proximalActivationPos}\includegraphics[width=0.42\textwidth]{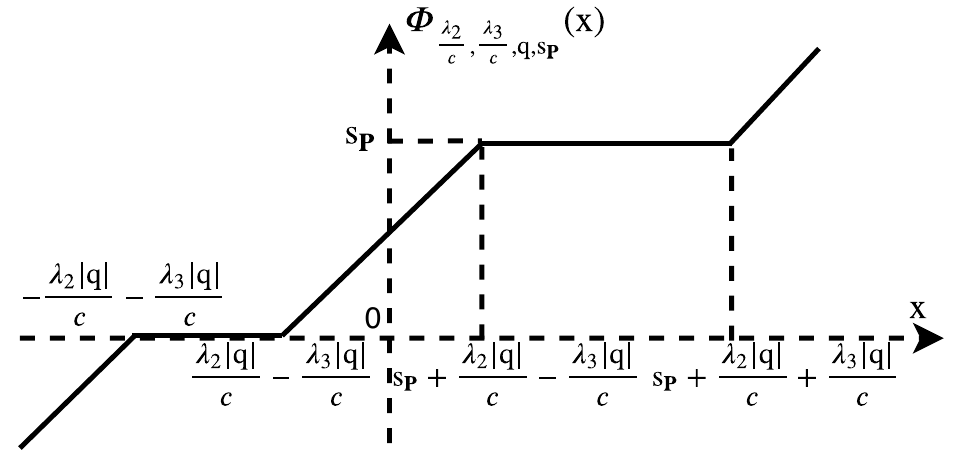}	}	
	\subfigure[refRPCA: $\varPhi_{\frac{\lambda_2}{c},\frac{\lambda_3}{c},\q,\s_{\bP}}(\mathrm{x})$ for $\s_{\bP}< 0$.]{\label{proximalActivationNeg}\includegraphics[width=0.42\textwidth]{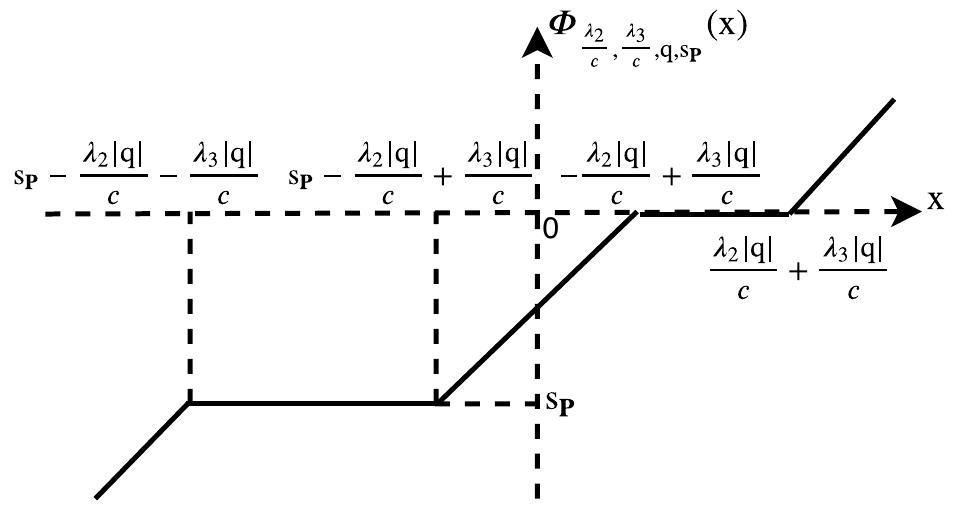}	}		
	\caption{The proximal operators of CORONA \cite{solomon2019deep} (a) vs refRPCA (b) and (c). The parameters $\lambda_2, \lambda_3$ and $c$ are learned globally. Note that (b) and (c) are drawn for given $\q$ and $\s_{\bP}$. The weight $\q$ allows for a different proximal operator for each entry of $\mathrm{x}$ due to the varying length of the multiple-threshold intervals. The reference $\s_{\bP}$ defines the position of the non-zero plateau each time the operator is evaluated.}
	\label{proximalActivation}
	\vspace{-8pt}
\end{figure} 
Since Problem \eqref{proximalOperatorS} is separable, it can be formulated element-wise. Let $\q$, $\s_{\bP}$, $\mathrm{x}$, $\uu$ denote each element of the corresponding $\bQ$, $\bS_{\bP}$, $\bX$, $\bU$. Then, 
\begin{align}\label{element-wise-proximalOperatorS}
\varPhi_{\frac{\lambda_2}{c},\frac{\lambda_3}{c},\q,\s_{\bP}}(\mathrm{x})=\arg\min\limits_{\uu}\Big\{&\frac{\lambda_2}{c}|\q\uu|+\frac{\lambda_3}{c}|\q(\uu-\s_{p})|\nonumber\\
&+ \frac{1}{2}(\uu-\mathrm{x})^{2} \Big\}.
\end{align}
The solution of \eqref{element-wise-proximalOperatorS} is derived in \cite{Luong2018Compressive} and is given as follows. For $\s_{\bP} \geq 0$:
\begin{align}
&\varPhi_{\frac{\lambda_2}{c},\frac{\lambda_3}{c},\q,\s_{\bP}}(\mathrm{x})=\nonumber\\
&\begin{cases}
\mathrm{x} - \frac{\lambda_2|\q|}{c} - \frac{\lambda_3|\q|}{c}, &\hspace{-5pt} \s_{\bP} + \frac{\lambda_2|\q|}{c}+ \frac{\lambda_3|\q|}{c} < \mathrm{x} < \infty \\
\s_{\bP}, &\hspace{-25pt} \s_{\bP} + \frac{\lambda_2|\q|}{c} - \frac{\lambda_3|\q|}{c} \leq \mathrm{x} \leq \s_{\bP} + \frac{\lambda_2|\q|}{c} + \frac{\lambda_3|\q|}{c} \\
\mathrm{x} - \frac{\lambda_2|\q|}{c} + \frac{\lambda_3|\q|}{c}, &\hspace{-5pt} \frac{\lambda_2|\q|}{c} - \frac{\lambda_3|\q|}{c} <\mathrm{x} < \s_{\bP} + \frac{\lambda_2|\q|}{c}
- \frac{\lambda_3|\q|}{c}\\
0, &\hspace{-5pt} -\frac{\lambda_2|\q|}{c}- \frac{\lambda_3|\q|}{c}\leq \mathrm{x} \leq \frac{\lambda_2|\q|}{c}- \frac{\lambda_3|\q|}{c}\\
\mathrm{x} + \frac{\lambda_2|\q|}{c} + \frac{\lambda_3|\q|}{c}, &\hspace{-5pt} -\infty < \mathrm{x} < -\frac{\lambda_2|\q|}{c} - \frac{\lambda_3|\q|}{c},\\
\end{cases}\label{reweighted-l1_positive}
\end{align}
and for $\s_{\bP} <0$:
\begin{align}
&\varPhi_{\frac{\lambda_2}{c},\frac{\lambda_3}{c}\q,\s_{\bP}}(\mathrm{x})=\nonumber\\
&\begin{cases}
\mathrm{x} - \frac{\lambda_2|\q|}{c} - \frac{\lambda_3|\q|}{c}, &\hspace{-5pt} \frac{\lambda_2|\q|}{c}+ \frac{\lambda_3|\q|}{c} < \mathrm{x} < \infty \\
0, &\hspace{-5pt} - \frac{\lambda_2|\q|}{c} +\frac{\lambda_3|\q|}{c} \leq \mathrm{x} \leq \frac{\lambda_2|\q|}{c} + \frac{\lambda_3|\q|}{c} \\
\mathrm{x} + \frac{\lambda_2|\q|}{c} - \frac{\lambda_3|\q|}{c} , & \hspace{-5pt}\s_{\bP} - \frac{\lambda_2|\q|}{c} +\frac{\lambda_3|\q|}{c} < \mathrm{x} < - \frac{\lambda_2|\q|}{c}
+ \frac{\lambda_3|\q|}{c}\\
\s_{\bP}, &\hspace{-25pt} \s_{\bP}-\frac{\lambda_2|\q|}{c}- \frac{\lambda_3|\q|}{c}\leq \mathrm{x} \leq \s_{\bP} - \frac{\lambda_2|\q|}{c} + \frac{\lambda_3|\q|}{c}\\
\mathrm{x} + \frac{\lambda_2|\q|}{c} + \frac{\lambda_3|\q|}{c}, &\hspace{-5pt} -\infty < \mathrm{x} <\s_{\bP} -\frac{\lambda_2|\q|}{c} - \frac{\lambda_3|\q|}{c}.\label{reweighted-l1_negative}\\
\end{cases}
\end{align}

The proposed refRPCA leads to the proximal operator $\varPhi_{\frac{\lambda_2}{c},\frac{\lambda_3}{c},\q,\s_{\bP}}(.)$ in \eqref{element-wise-proximalOperatorS}, which replaces $\varPhi_{\frac{\lambda_2}{c}}(.)$ \eqref{SProximal} in CORONA \cite{solomon2019deep}. The difference is illustrated in Fig. \ref{proximalActivation}: Fig. \ref{proximalCORONA} shows the proximal operator for CORONA; Figs.~\ref{proximalActivationPos}~and~\ref{proximalActivationNeg} depict the generic form of the proximal operator for $\s_{\bP} \geq 0$ [Fig. \ref{proximalActivationPos}] and $\s_{\bP}< 0$ [Fig. \ref{proximalActivationNeg}]. Observe that the proximal function for CORONA [Fig. \ref{proximalCORONA}] has a single threshold, while, for refRPCA [Fig. \ref{proximalActivationPos} and Fig. \ref{proximalActivationNeg}], different values of $\q$ lead to different shapes of the proximal functions $\varPhi_{\frac{\lambda_2}{c},\frac{\lambda_3}{c},\q,\s_{\bP}}(.)$ for each entry of the input, due to the varying length of the multiple-threshold intervals.

The algorithm for solving Problem \eqref{refRPCA} is summarized in Algorithm \ref{refRPCA-algorithm}, where $\varPhi_{\frac{\lambda_2}{c},\frac{\lambda_3}{c},\bq,\bS_{\bP}}(.)$ in Line \ref{alg:proximalOperator} is given by \eqref{reweighted-l1_positive} and \eqref{reweighted-l1_negative}. As shown in Algorithm \ref{refRPCA-algorithm}, we need to properly tune several parameters: $\bP$, $\bQ$, $\lambda_1$, $\lambda_2$, $\lambda_3$, which play significant roles in the singular value thresholding and proximal operators [see Lines \ref{alg:SVTOperator} and \ref{alg:proximalOperator}]. 
\begin{algorithm}[t!]
	\textbf{Input:} $\bM$, $\bP$, $\bQ$, $\bH_1$, $\bH_2$, $\lambda_1$, $\lambda_2$, $\lambda_3$, $c$, the maximum number of iterations $d$.\\
	\textbf{Output:} $\widehat{\bL}$, $\widehat{\bS}$.\\
	\For{k = 1 to d}{
		$\widetilde{\bL}^{(k)} = \Big(\bI-\frac{1}{c}\bH_1^T\bH_1\Big)\bL^{(k)}-\bH_1^T\bH_2\bS^{(k)}+\bH_1^T\bM$\\
		$\widetilde{\bS}^{(k)} = \Big(\bI-\frac{1}{c}\bH_2^T\bH_2\Big)\bS^{(k)}-\bH_2^T\bH_1\bL^{(k)}+\bH_2^T\bM$\\	
		$\bL^{(k+1)} = \varGamma_{\frac{\lambda_1}{c}}\Big(\widetilde{\bL}^{(k)}\Big)$\label{alg:SVTOperator}\\
		$\bS_{\bP} = [\bs^{(k)}_1,\bP\bs^{(k)}_1,\cdots,\bP\bs^{(k)}_{m-1}]$\\
		$\bS^{(k+1)} = \varPhi_{\frac{\lambda_2}{c},\frac{\lambda_3}{c},\bq,\bS_{\bP}}\Big(\widetilde{\bS}^{(k)}\Big)$\label{alg:proximalOperator}\\
	}
	\Return $\widehat{\bL}=\bL^{(k+1)}$, $\widehat{\bS}=\bS^{(k+1)}$.
	\caption{The proposed refRPCA algorithm for foreground-background separation.}
	\label{refRPCA-algorithm}
	\end{algorithm}


\subsection{The refRPCA-Net Architecture}

\begin{figure}[t!]
	\centering 
	\includegraphics[width=0.4\textwidth]{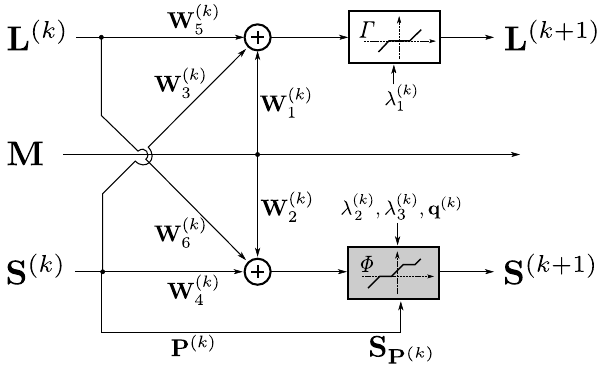}			
	\caption{The proposed refRPCA-Net architecture.}
	\label{refRPCA-Net}
\end{figure}

We now propose to unroll the iterations of the refRPCA algorithm into an $d$-layer network, coined \textit{deep-unfolded reference-based RPCA network} (refRPCA-Net). The $k^{\mathrm{th}}$ iteration in Algorithm \ref{refRPCA-algorithm} corresponds to the $k^{\mathrm{th}}$ layer in refRPCA-Net, which is illustrated in Fig.~\ref{refRPCA-Net}. At each layer $k$, the low-rank component $\bL^{(k+1)}$ is updated as in \eqref{LLayer}, while $\bS^{(k+1)}$ is updated as
	\begin{align}
	\hspace{-3pt}\bS^{(k+1)}\hspace{-3pt}= \varPhi_{\lambda_2^{(k)},\lambda_3^{(k)},\bq^{(k)},\bS_{\bP^{(k)}}}\Big\{&\bW_2^{(k)}*\bM+\bW_4^{(k)}*\bS^{(k)}\nonumber\\
	&+\bW_6^{(k)}*\bL^{(k)}\Big\},\label{refSLayer}
	\end{align}
	where the parameters $\lambda_1^{(k)}, \lambda_2^{(k)},\lambda_3^{(k)}, \bq^{(k)}, \bP^{(k)}$ and the parameters of the convolutional layers $\bW_1^{(k)},\cdots,\bW_6^{(k)}$ are learned per layer during training. 
	
	The key difference with CORONA \cite{solomon2019deep} is the proximal operator (a.k.a., activation function) $\varPhi$ [see the block highlighted in gray color in Fig.~\ref{refRPCA-Net}]. $\varPhi$ incorporates the reference $\bS_{\bP^{(k)}}$ via the matrix $\bP^{(k)}$, defining by $\bS_{\bP^{(k)}} = [\bs^{(k)}_1,\bP^{(k)}\bs^{(k)}_1,\cdots,\bP^{(k)}\bs^{(k)}_{m-1}]$. Furthermore, the different learnable values of $\q_i\in \bq^{(k)}$ lead to different realizations of the activation functions $\varPhi_{\lambda_2^{(k)},\lambda_3^{(k)},\q_i,\s_{\bP^{(k)}}}(.)$ for each neuron [see Figs. \ref{proximalActivationPos} and \ref{proximalActivationNeg}]. This increases the adaptivity and expressivity of refRPCA-Net. We note that the low-rank component $\bL^{(k)}$ is updated in a similar manner as in CORONA. However, after each layer, the updated $\bS^{(k)}$ becomes one of the inputs for updating $\bL^{(k+1)}$ [see \eqref{LLayer}]. Therefore, an improvement in the estimation of $\bS$ leads to an improvement in the estimation of $\bL$ and vice versa.

	\section{Experiments}
	\label{sec:experiment}
	
	In this section we assess the performance of the proposed refRPCA-Net versus CORONA~\cite{solomon2019deep} in the task of video foreground-background separation. We consider the moving MNIST dataset~\cite{SrivastavaICML15} for our experiments, which contains 10,000 sequences of moving digits, each 20 frames long. We resize each frame in the dataset to a resolution of $32\times 32$ pixels, thereby having $m=20$ and $n=1024$ according to our notation (see~Section~\ref{sec:proposedNetwork}). We add a synthetic low-rank background to each sequence that is generated as in~\cite{Luong2018Compressive}; namely, we generate $\mathbf{L} \doteq \mathbf{UV}^T$, with $\mathbf{U} \in \mathbb{R}^{n\times r}$ and $\mathbf{V} \in \mathbb{R}^{m\times r}$ sampled from a standard Gaussian distribution and the rank set to $r=5$. The dataset is then split into 8000, 1000, and 1000 samples respectively for training, validation, and testing. The pixel intensities are normalized to the unit range before being fed to the network.
	The two models are trained using the Adam optimizer with a learning rate of $10^{-3}$, a batch size of 200 and during 50 epochs by minimizing the following compound mean square error (MSE) loss:
	\begin{equation}
	  \mathcal{L}(\varTheta) = \frac{1}{2N} \sum_{i=1}^N \|\mathbf{L}_i-\widehat{\mathbf{L}}_i\|_F^2 + \frac{1}{2N} \sum_{i=1}^N\|\mathbf{S}_i - \widehat{\mathbf{S}}_i\|_F^2,
	\end{equation}
	where $\varTheta=\Big\{\bW_1^{(k)},\cdots,\bW_6^{(k)}, \lambda_1^{(k)}, \lambda_2^{(k)}, \lambda_3^{(k)}, \bq^{(k)}, \bP^{(k)}\Big\}_{k=1}^{d}$ are the learnt parameters, $\{\bS_i,\bL_i\}_{i=1}^{N}$ are the ground-truth and $\{\widehat{\bS}_i,\widehat{\bL}_i\}_{i=1}^{N}$ the foreground and background components predicted by the network (with $\bS_i,\bL_i,\widehat{\bS}_i,\widehat{\bL}_i \in \mathbb{R}^{1024\times 20}$), and $N$ is the number of training samples in the dataset. Both networks are trained with different number of layers, that is, from 1 to 10 layers. The remaining network configurations, including the convolutional kernel sizes and the initialization of the weights, are kept identical to what is reported in~\cite{CohenICASSP19} to ensure a fair comparison. 
		\begin{figure}[t!]
	\centering
	\subfigure[Average MSE.]{\label{M_mse}\includegraphics[width=0.4\textwidth]{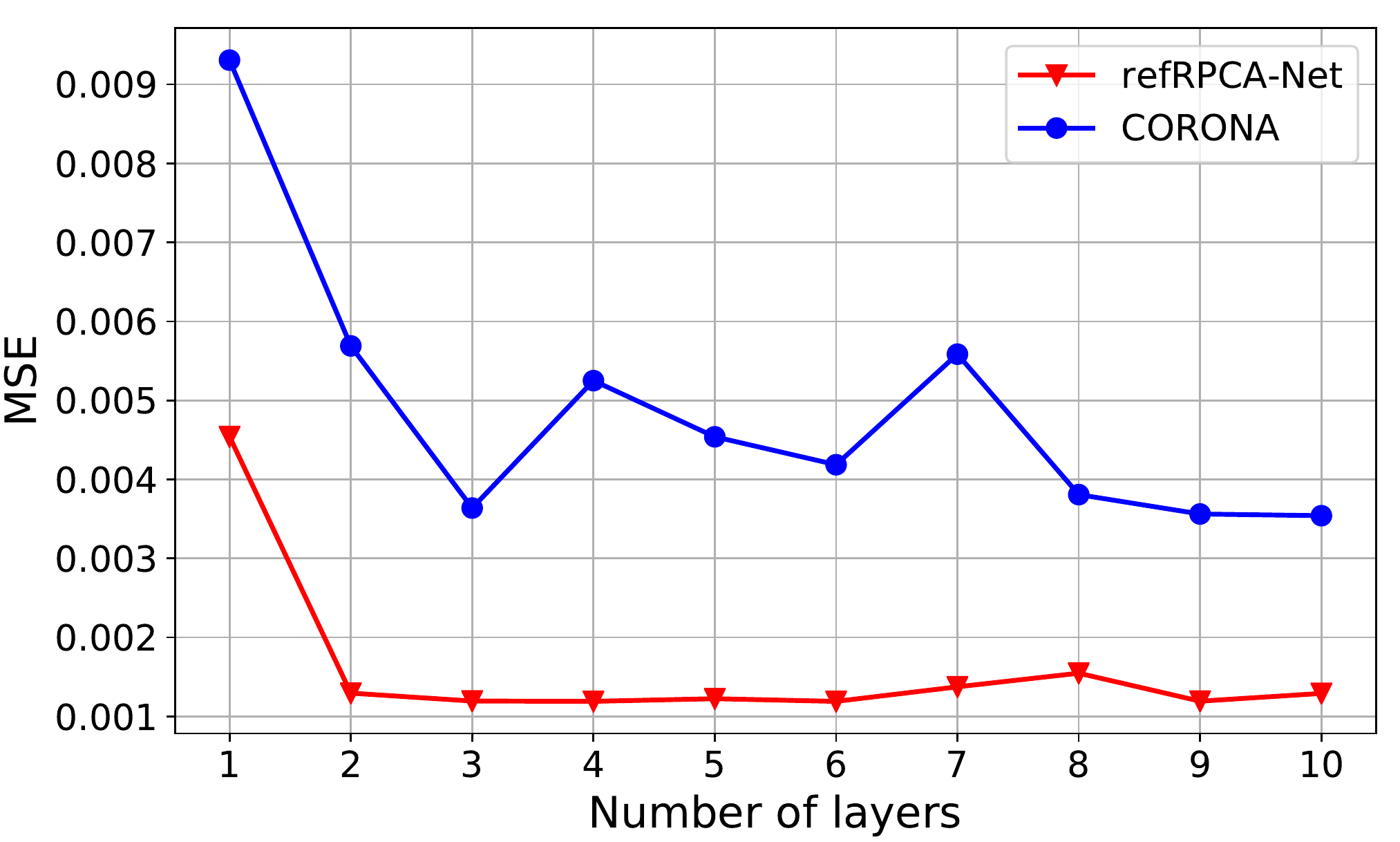}}
	\subfigure[MSE for $\bL$ and $\bS$.]{\label{LS_mse}\includegraphics[width=0.40\textwidth]{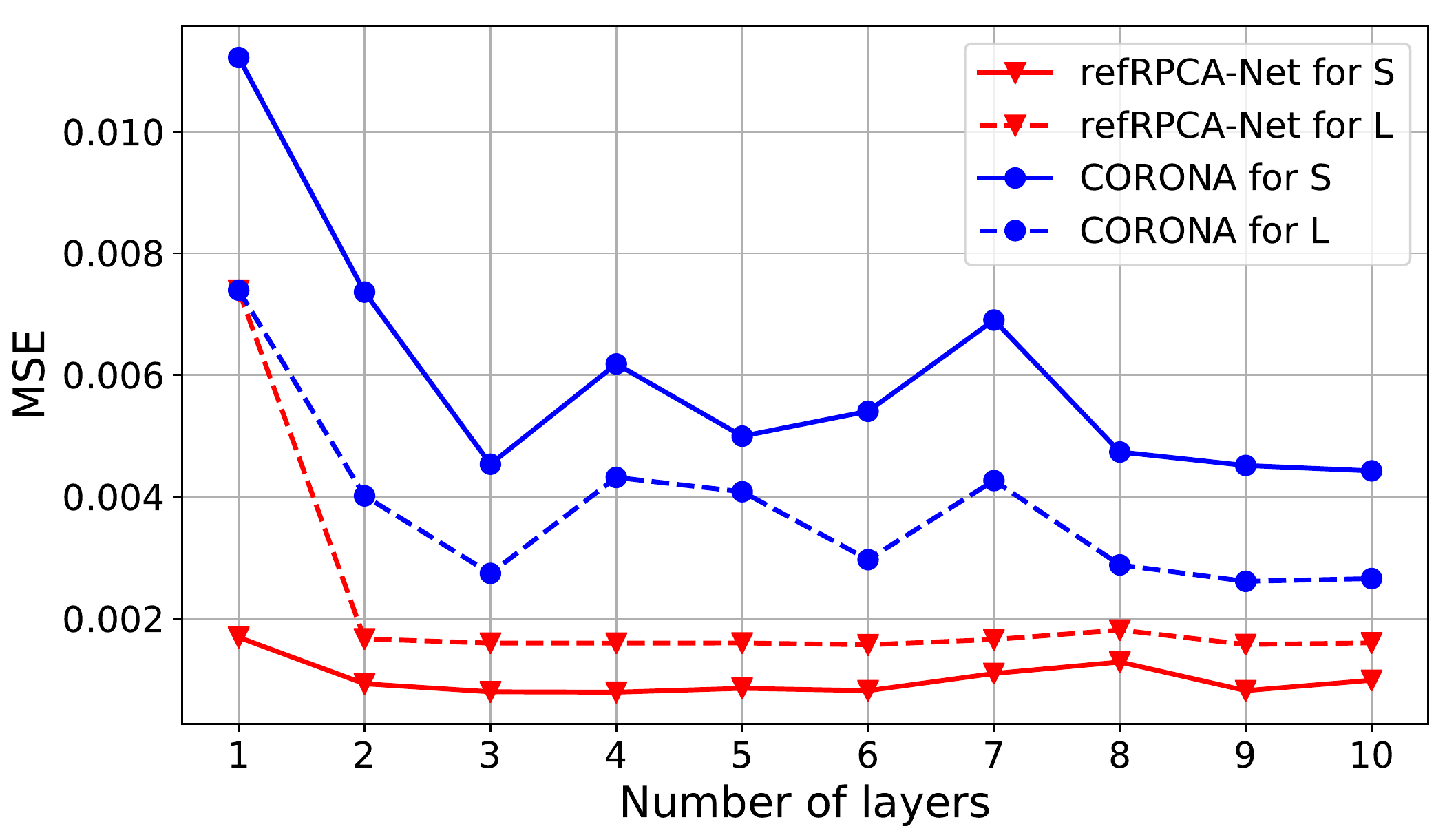}}
	\caption{Average Mean Square Error (MSE) vs. the number of layers for the proposed refRPCA-Net and CORONA~\cite{CohenICASSP19} on video separation.}
	\end{figure} 
	
\begin{figure}[t!]
	\centering
	\subfigure[1 layer.]{\label{refRPCA_1layer_bg}\includegraphics[width=0.075\textwidth]{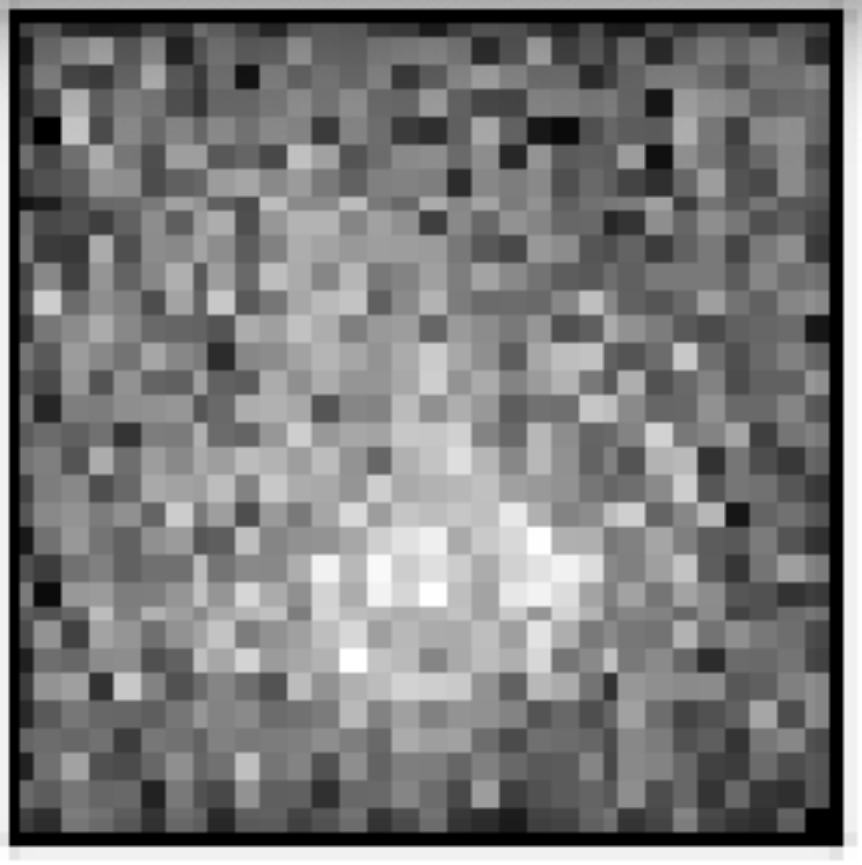}}
	\subfigure[1 layer.]{\label{refRPCA_1layer_fg}\includegraphics[width=0.075\textwidth]{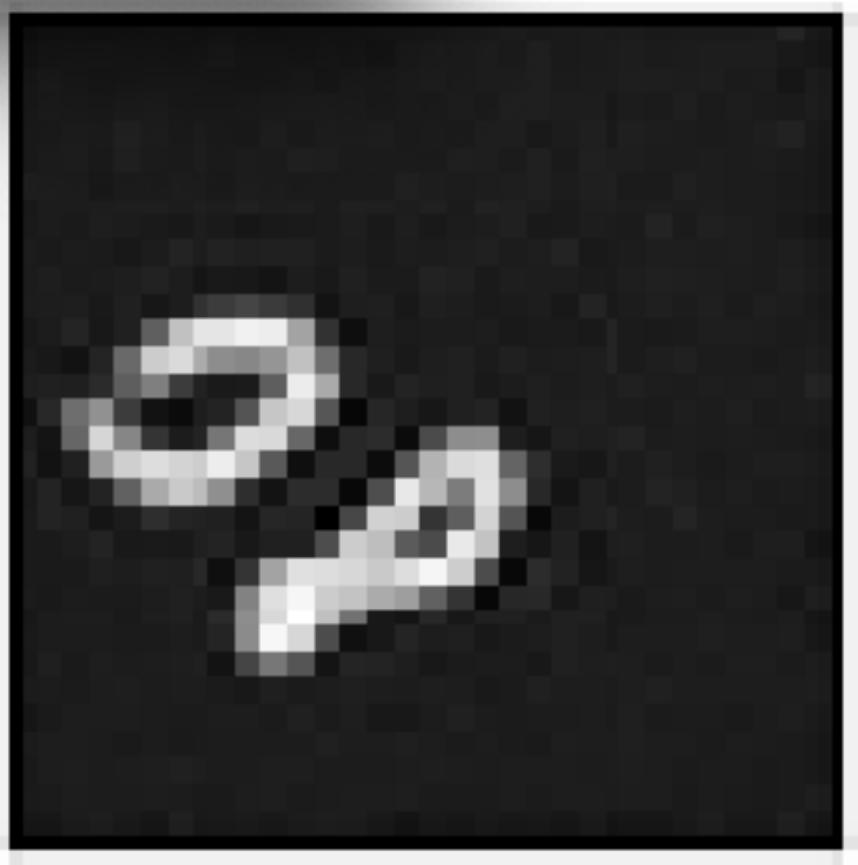}}
	\subfigure[\hspace{-2pt}2 layers.]{\label{refRPCA_2layer_bg}\includegraphics[width=0.075\textwidth]{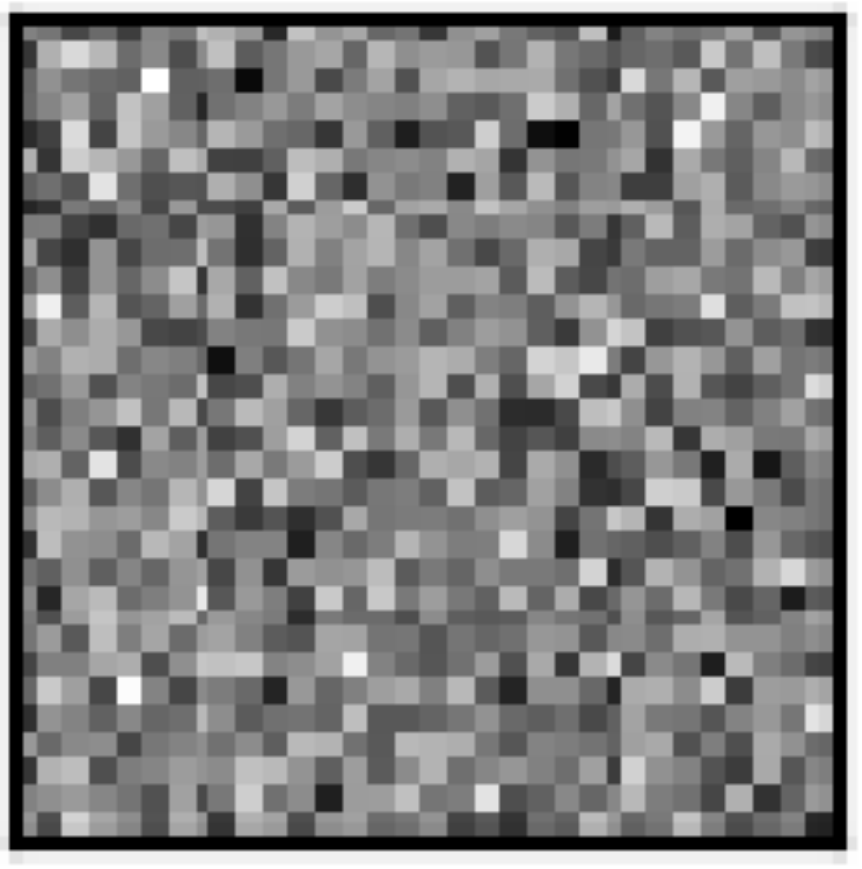}}
	\subfigure[\hspace{-2pt}2 layers.]{\label{refRPCA_2layer_fg}\includegraphics[width=0.075\textwidth]{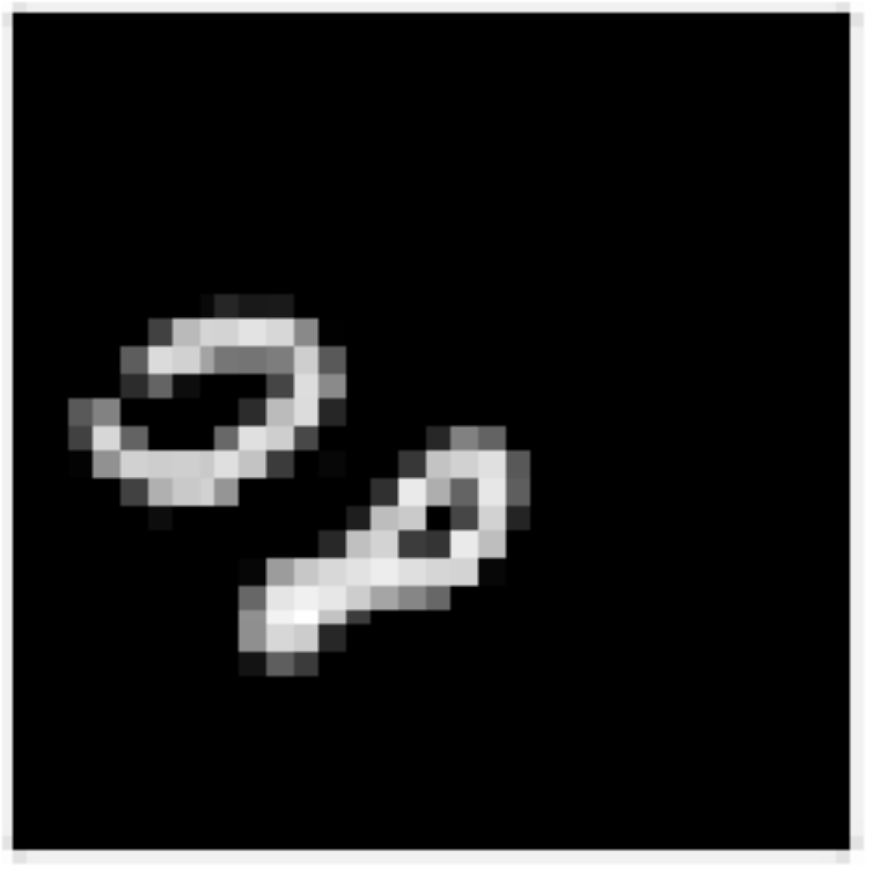}}
	\subfigure[\hspace{-3pt}10 layers.\hspace{-3pt}]{\label{refRPCA_10layer_bg}\includegraphics[width=0.075\textwidth]{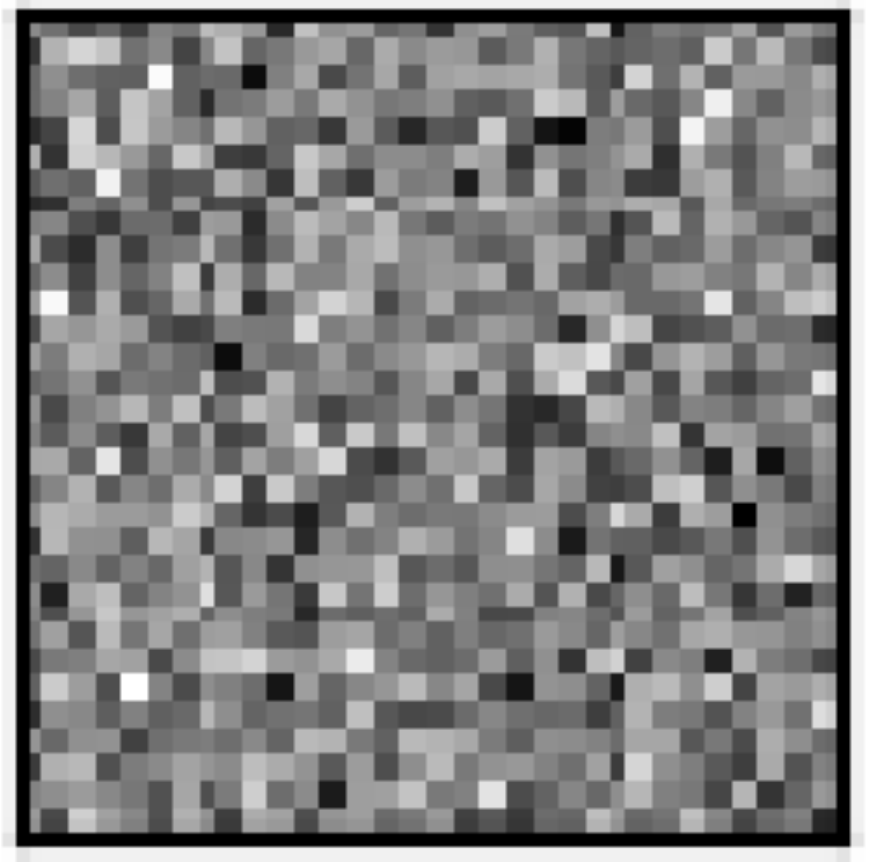}}
	\subfigure[\hspace{-3pt}10 layers.\hspace{-2pt}]{\label{refRPCA_10layer_fg}\includegraphics[width=0.075\textwidth]{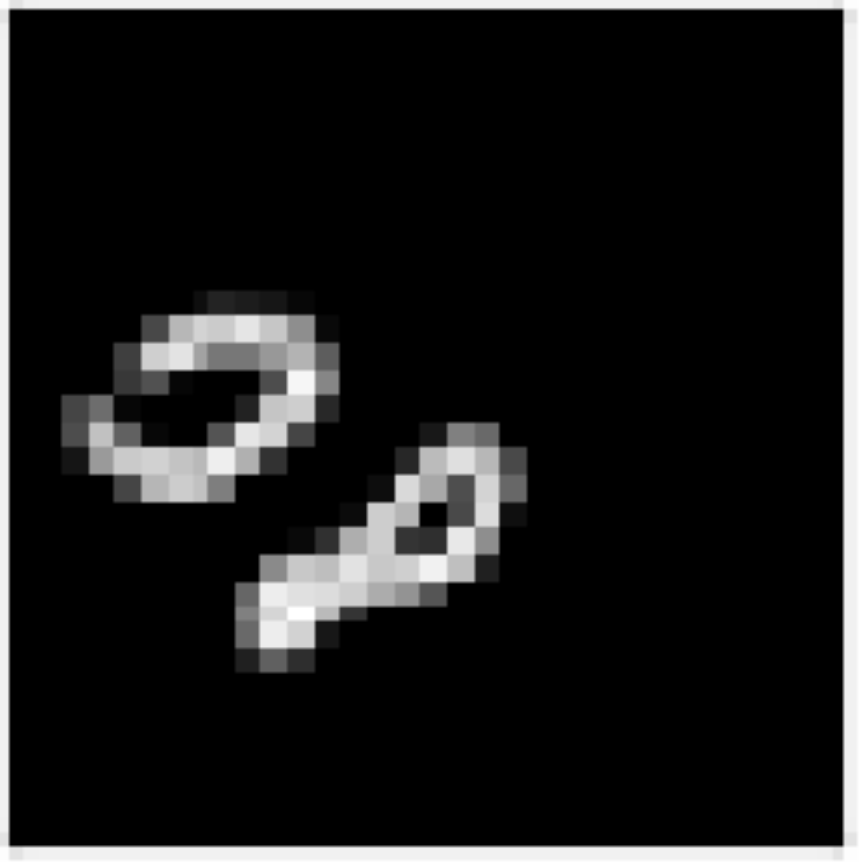}}	
	\vspace{-5pt}
	\caption{Visual results for refRPCA-Net for different number of layers: (a), (c), (e) Background frames and (b), (d), (f) Foreground frames.}\label{refRPCA_images}
	\end{figure}

\begin{figure}[t!]
	\centering
	\subfigure[\hspace{-2pt}1 layer.]{\label{corona_1layer_bg}\includegraphics[width=0.075\textwidth]{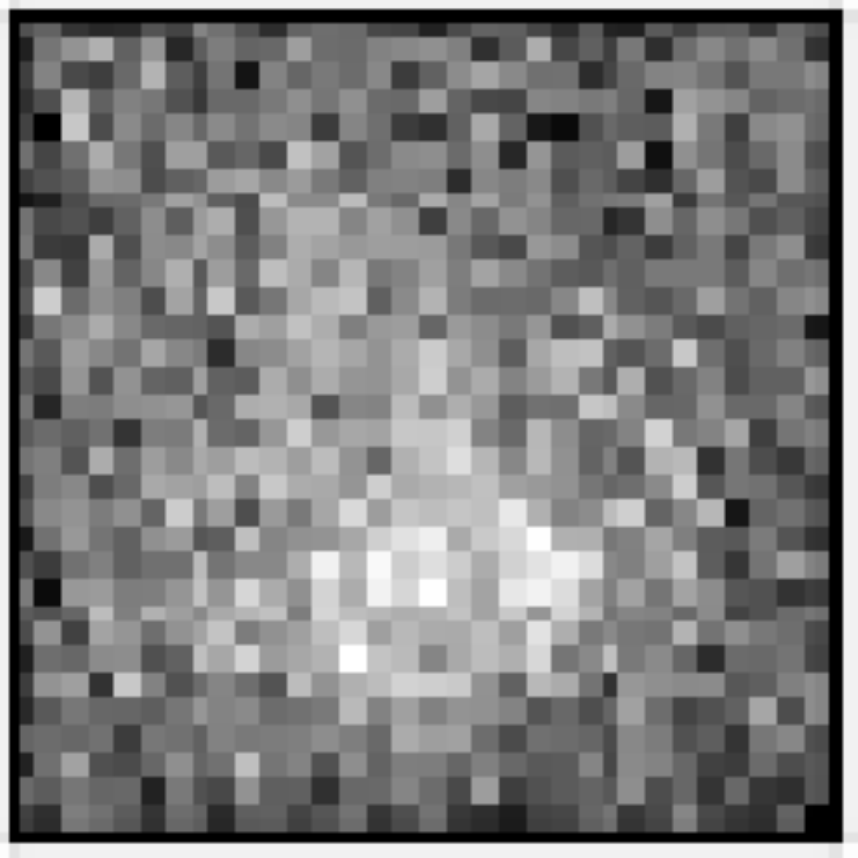}}
	\subfigure[\hspace{-2pt}1 layer.]{\label{corona_1layer_fg}\includegraphics[width=0.075\textwidth]{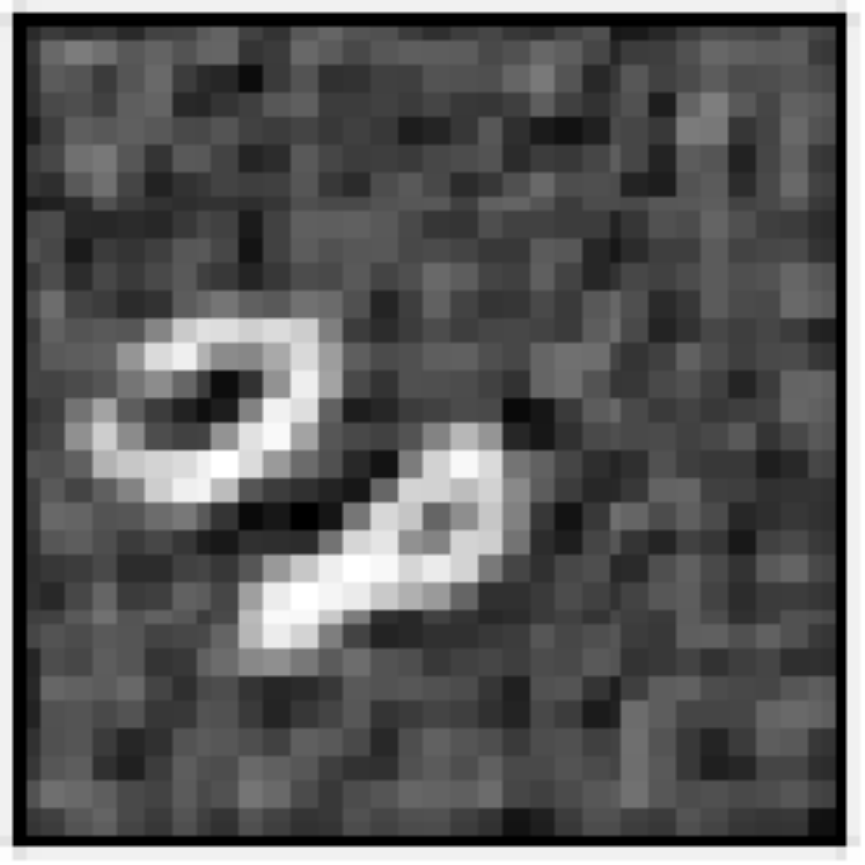}}
	\subfigure[\hspace{-2pt}2 layers.]{\label{corona_2layer_bg}\includegraphics[width=0.075\textwidth]{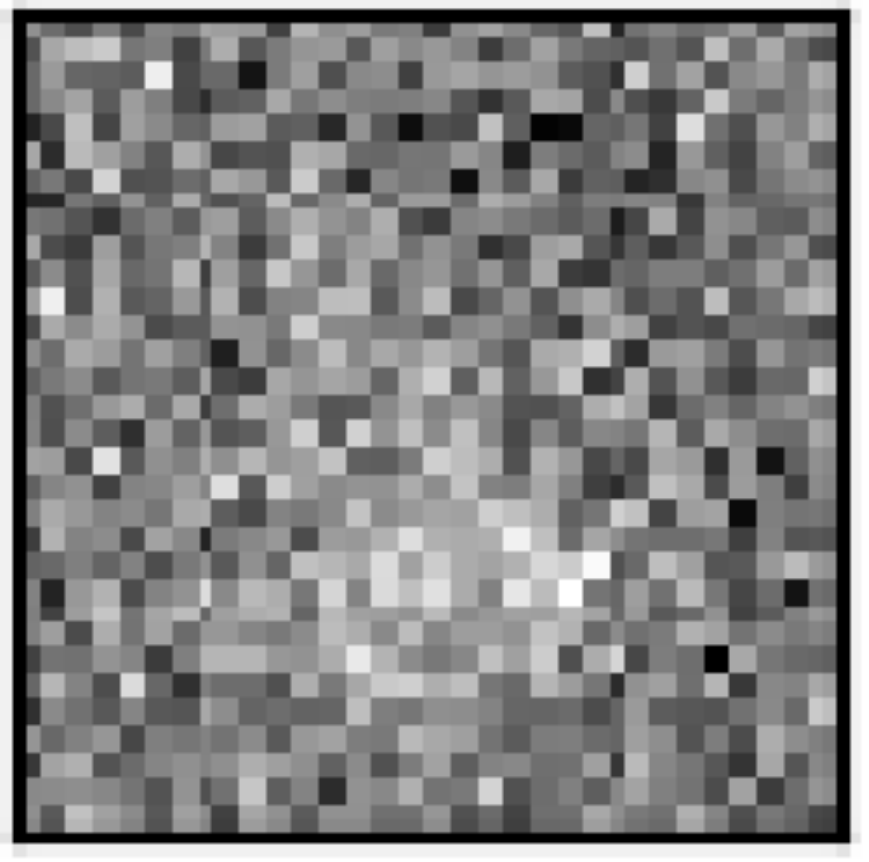}}
	\subfigure[\hspace{-2pt}2 layers.]{\label{corona_2layer_fg}\includegraphics[width=0.075\textwidth]{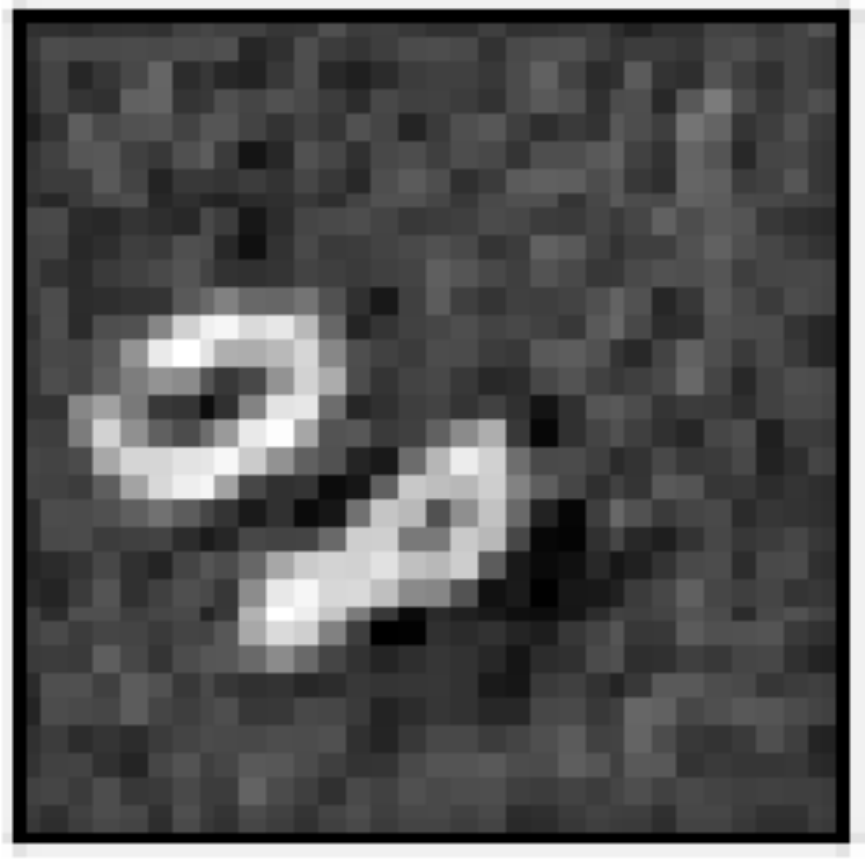}}
	\subfigure[\hspace{-3pt}10 layers.\hspace{-3pt}]{\label{corona_10layer_bg}\includegraphics[width=0.075\textwidth]{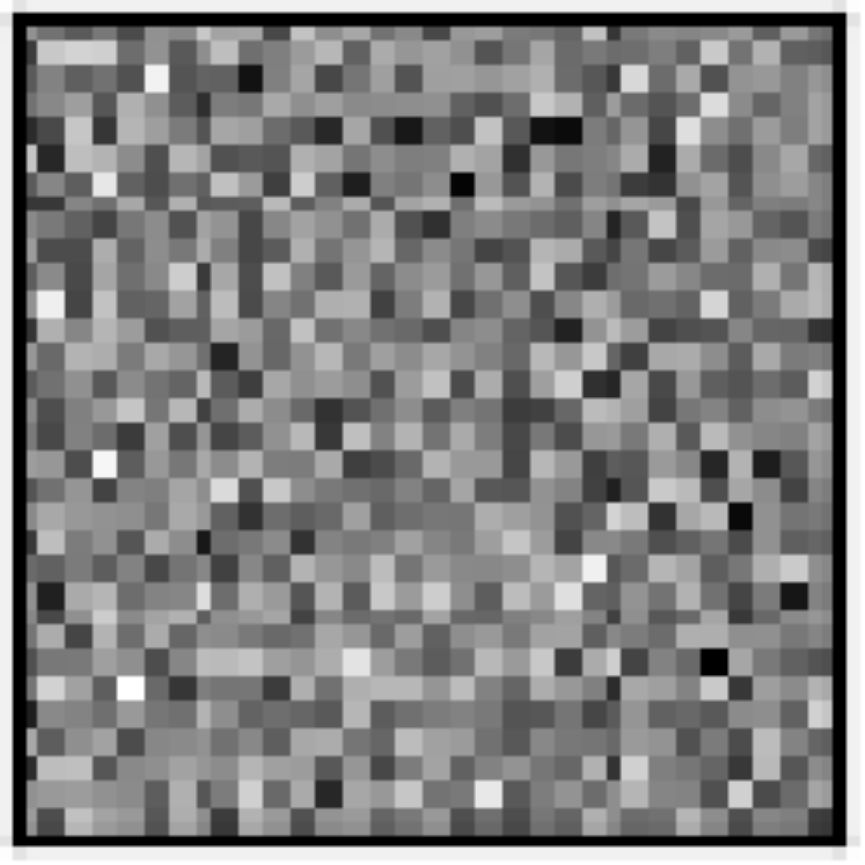}}
	\subfigure[\hspace{-3pt}10 layers.\hspace{-2pt}]{\label{corona_10layer_fg}\includegraphics[width=0.075\textwidth]{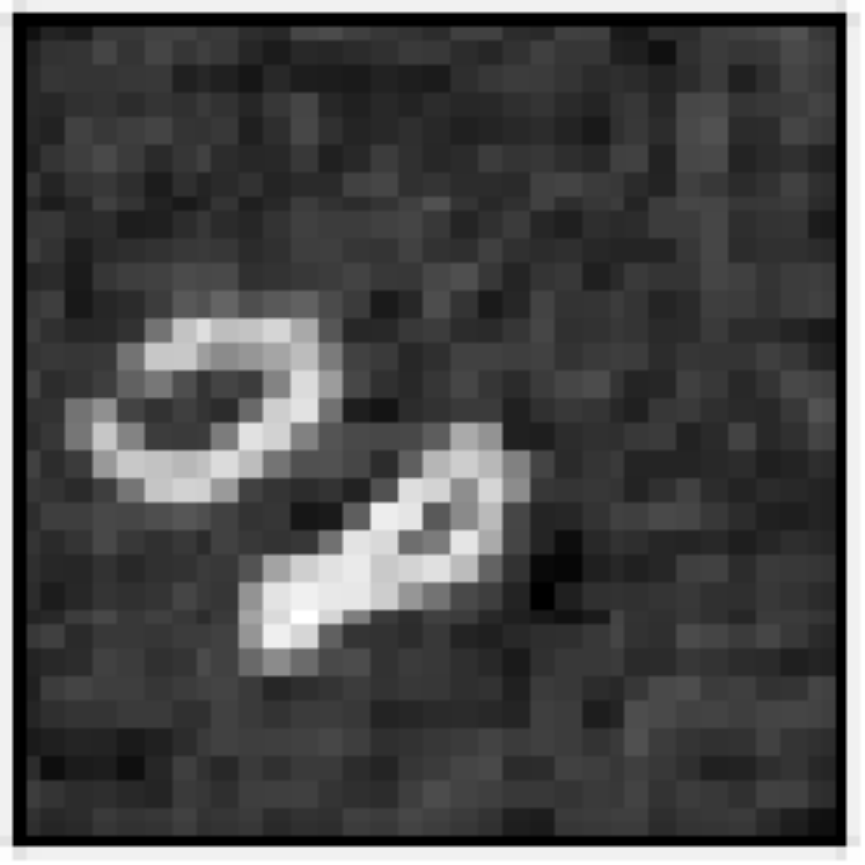}}	
	\vspace{-5pt}
	\caption{Visual results for CORONA for different number of layers: (a), (c), (e) Background frames and (b), (d), (f) Foreground frames.}\label{CORONA_images}
	\end{figure}

	We plot the reconstruction mean squared-error (MSE)---averaged over the sequences in the validation set---versus the number of layers for each model; specifically, Fig.~\ref{M_mse} reports the average MSE of both components whereas Fig.~\ref{LS_mse} reports the average MSE for the low-rank and sparse components separately. We observe that the proposed refRPCA-Net outperforms CORONA. Similar to what has been reported in~\cite{solomon2019deep}, for both networks, increasing the number of layers above a certain number does not lead to a significant performance improvement, while inducing higher complexity in training and inference. The proposed network, however, achieves faster convergence and delivers more stable performance to the number of layers than CORONA. Additionally, a foreground-background separation example is displayed in Fig.~\ref{refRPCA_images} for refRPCA-Net and Fig.~\ref{CORONA_images} for CORONA, for the 1, 2 and 10 layers configurations. Again, we observe that refRPCA-Net leads to better estimates of the sparse and low-rank components compared to CORONA.
	\begin{figure}[t!]
	\centering
	\vspace{-5pt}
	\subfigure[]{\label{original}\includegraphics[width=0.075\textwidth]{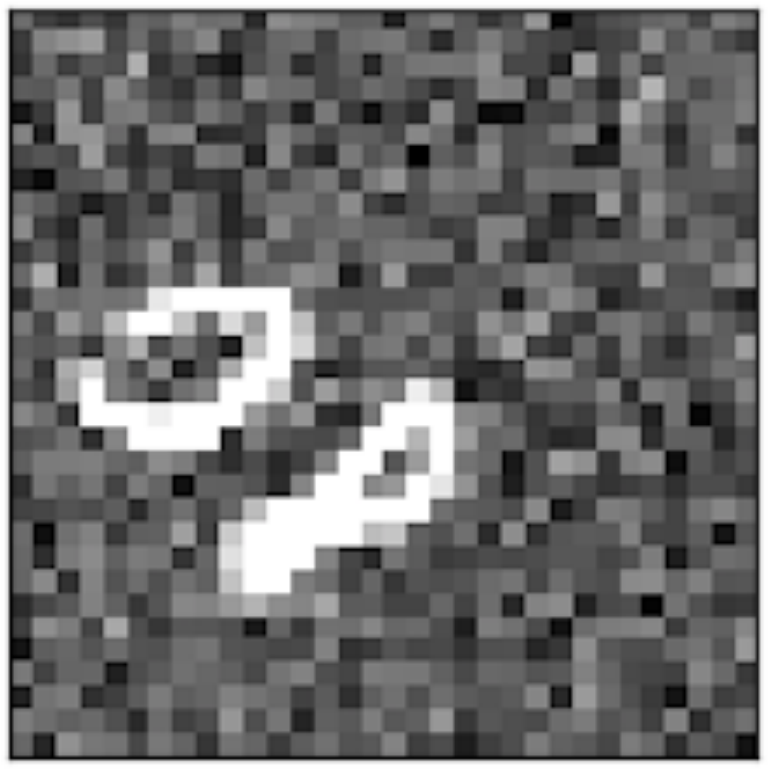}}
	\hspace{18pt}\subfigure[]{\label{background}\includegraphics[width=0.075\textwidth]{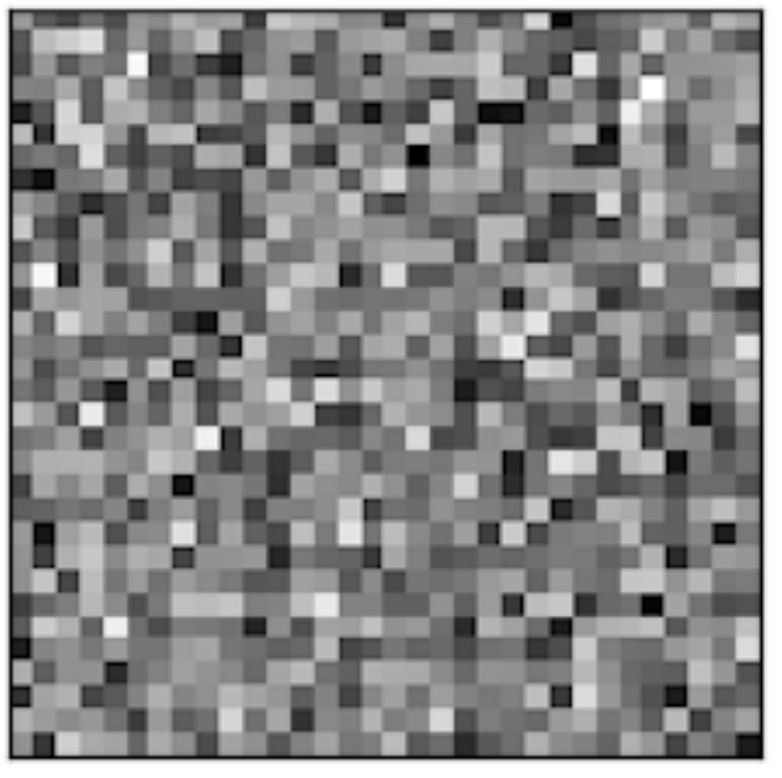}}
		\hspace{18pt}\subfigure[]{\label{foreground}\includegraphics[width=0.075\textwidth]{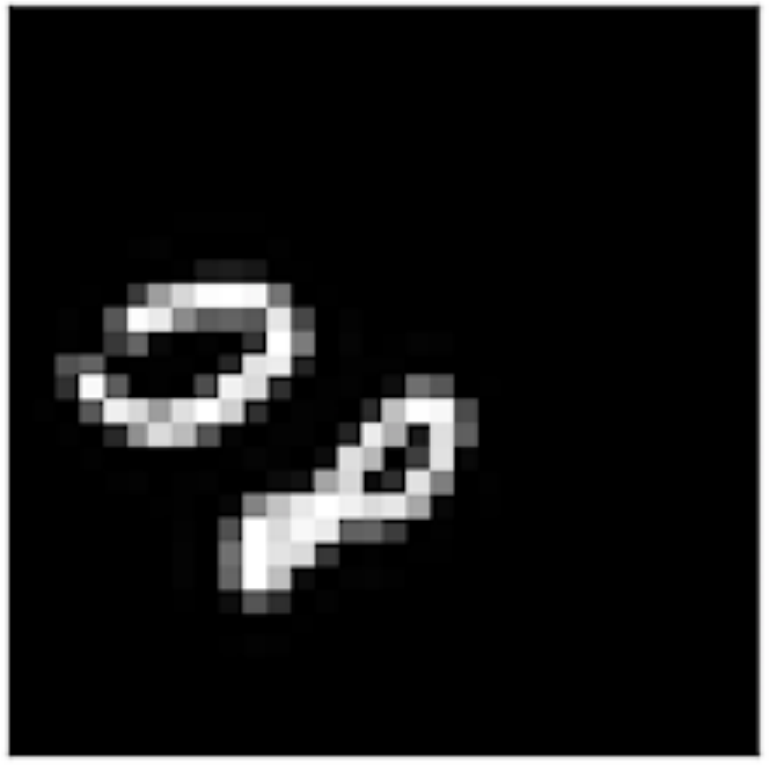}}
		\vspace{-5pt}
	\caption{(a) Original frame,  Ground-truth (b) background and (c) foreground.}\label{original_images}
	\end{figure} 	
	
  The performance gain of refRPCA-Net compared to CORONA is higher for the sparse component because the method directly leverages the reference $\mathbf{S}_{\mathbf{P}^{(k)}}$ through the reweighted activation function. Furthermore, while the two networks use the same update rule for the low-rank component, the gain in the sparse component estimation in refRPCA-Net indirectly leads to an improved estimation of the low-rank component as well. This gain is notable when the number of network layers is higher than 1. When the number of layers is set to 1 (corresponding to one iteration of the method), the enhanced estimation of the sparse component is not exploited to improve the reconstruction of the low-rank component.

	

	\section{Conclusion}
   We proposed a deep-unfolded reference-based RPCA network for the task of video foreground-background separation. Our refRPCA-Net architecture captures the correlation between the sparse components of consecutive video frames by unfolding an algorithm for solving reweighted $\ell_1$-$\ell_1$ minimization. Our design leads to a different proximal operator (activation function) adaptively learnt per neuron. Our experiments using the moving MNIST dataset show that our network outperforms the state-of-the-art CORONA network consistently for the reconstruction of both the foreground and background components. 
	\label{sec:conclusion}

	
	\bibliographystyle{IEEEbib}
	\bibliography{refRPCA-Net_paper.bib}
\end{document}